\documentclass[runningheads]{llncs}
\usepackage{graphicx}
\usepackage{amsmath,amssymb} % define this before the line numbering.
\usepackage{color}
\usepackage{multirow}
\usepackage{amsmath}
\usepackage{algorithm}
\usepackage{algpseudocode}
\usepackage{algorithmicx}
\usepackage{caption}
\usepackage{mmstyles}

\graphicspath{{figures/}}
\DeclareGraphicsExtensions{.pdf}

\algnewcommand\algorithmicinput{\textbf{Input:}}
\algnewcommand\INPUT{\item[\algorithmicinput]}
\algnewcommand\algorithmicoutput{\textbf{Output:}}
\algnewcommand\OUTPUT{\item[\algorithmicoutput]}

\begin{document}
\title{Consensus-Driven Propagation in \\ Massive Unlabeled Data for Face Recognition}
% Replace with your title

\titlerunning{Consensus-Driven Propagation}
% Replace with a meaningful short version of your title

\author{Xiaohang Zhan\inst{1}\orcidID{0000-0003-2136-7592} 
	\and Ziwei Liu\inst{1}\orcidID{0000-0002-4220-5958}
	\and \\Junjie Yan\inst{2} 
	\and Dahua Lin\inst{1}\orcidID{0000-0002-8865-7896} 
	\and \\Chen Change Loy\inst{3}\orcidID{0000-0001-5345-1591}} % First name: Chen Change, Last name: Loy

\authorrunning{X. Zhan, Z. Liu, D. Lin, and C. C. Loy}
% Replace with shorter version of the author list. If there are more authors than fits a line, please use A. Author et al.

\institute{CUHK - SenseTime Joint Lab, The Chinese University of Hong Kong\\
	\email{\{zx017, zwliu, dhlin\}@ie.cuhk.edu.hk} \\ \and 
 	SenseTime Group Limited \hspace{1cm} $^\mathrm{3}$ Nanyang Technological University \\
    \hspace{-1cm} \email{yanjunjie@sensetime.com} \hspace{2.5cm} \email{ccloy@ieee.org} 
}

\maketitle
\setcounter{footnote}{0}

\begin{abstract}

Face recognition has witnessed great progress in recent years, mainly attributed to the high-capacity model designed and the abundant labeled data collected.
However, it becomes more and more prohibitive to scale up the current million-level identity annotations.
In this work, we show that unlabeled face data can be as effective as the labeled ones. 
Here, we consider a setting closely mimicking the real-world scenario, where the unlabeled data are collected from unconstrained environments and their identities are exclusive from the labeled ones.
Our main insight is that although the class information is not available, we can still faithfully approximate these semantic relationships by constructing a relational graph in a bottom-up manner.
We propose Consensus-Driven Propagation (CDP) to tackle this challenging problem with two modules, the ``committee'' and the ``mediator'', which select positive face pairs robustly by carefully aggregating multi-view information.  
Extensive experiments validate the effectiveness of both modules to discard outliers and mine hard positives.
With CDP, we achieve a compelling accuracy of 78.18\% on MegaFace identification challenge by using only 9\% of the labels, comparing to 61.78\% when no unlabeled data are used and 78.52\% when all labels are employed.

%\keywords{Consensus-Driven Propagation; Massive Unlabeled Data; Face Recognition}

\end{abstract}

\section{Introduction}
\label{sec:introduction}

Modern face recognition system mainly relies on the power of high-capacity deep neural network coupled with massive annotated data for learning effective face representations~\cite{sun2014deep,liu2015deep,schroff2015facenet,wen2016discriminative,huang2018deep,cao2018pose,zhang2018accelerated}.
From CelebFaces~\cite{sun2014deepnips} ($200K$ images) to MegaFace~\cite{kemelmacher2016megaface} ($4.7M$ images) and MS-Celeb-1M~\cite{guo2016ms} ($10M$ images), face databases of increasingly larger scale are collected and labeled.
Though impressive results have been achieved, we are now trapped in a dilemma where there are hundreds of thousands manually labeling hours consumed behind each percentage of accuracy gains.
To make things worse, it becomes harder and harder to scale up the current annotation size to even more identities. 
In reality, nearly all existing large-scale face databases suffer from a certain level of annotation noises~\cite{chen2018the}; it leads us to question how reliable human annotation would be.
%In large-scale face recognition, computers already perform on par with human performance~\cite{lu2015surpassing}

To alleviate the aforementioned challenges, we shift the focus from obtaining more manually labels to leveraging more unlabeled data.
Unlike large-scale identity annotations, unlabeled face images are extremely easy to obtain. 
For example, using a web crawler facilitated by an off-the-shelf face detector would produce abundant in-the-wild face images or videos~\cite{sohn2017unsupervised}.
Now the critical question becomes how to leverage the huge existing unlabeled data to boost the performance of large-scale face recognition.
This problem is reminiscent of the conventional semi-supervised learning (SSL)~\cite{zhu2005semi}, but significantly differs from SSL in two aspects:
First, the unlabeled data are collected from unconstrained environments, where pose, illumination, occlusion variations are extremely large. 
It is non-trivial to reliably compute the similarity between different unlabeled samples in this in-the-wild scenario. 
Second, there is usually no identity overlapping between the collected unlabeled data and the existing labeled data.
Thus, the popular label propagation paradigm~\cite{zhu2002learning} is no longer feasible here. 
%Fig.~\ref{fig:intro}(a) and \ref{fig:intro}(b) illustrate the role of unlabeled data in conventional semi-supervised learning and in real-world face recognition.

%\begin{figure}[t]
%\centering
%\includegraphics[width=0.82\linewidth]{intro}
%%\vskip -0.2cm
%\caption{\small{(a) and (b) show the different roles of unlabeled data in conventional semi-supervised learning and face recognition. Colored nodes represent labeled samples and gray nodes stand for unlabeled samples. For face recognition, unlabeled data and labeled data typically have no shared categories. (c) shows the inner structure in unlabeled data. With a single model, the structure is usually disordered and indistinguishable. (d) shows our consensus-driven graph composed of the pairs selected by the ``committee'' and the ``mediator''. See Section~\ref{sec:cdp} for details.}}
%\label{fig:intro}
%%\vspace{-0.5cm}
%\end{figure}

In this work, we study this challenging yet meaningful semi-supervised face recognition problem, which can be formally described as follows. 
In addition to some labeled data with known face identities, we also have access to a massive number of in-the-wild unlabeled samples whose identities are exclusive from the labeled ones.
Our goal is to maximize the utility of the unlabeled data so that the final performance can closely match the performance when all the samples are labeled. 
One key insight here is that although unlabeled data do not provide us with the straightforward semantic classes, its inner structure, which can be represented by a graph, actually reflects the distribution of high-dimensional face representations.
The idea of using a graph to reflect structures is also adopted in cross-task tuning~\cite{zhan2017mix}.
With the graph, we can sample instances and their relations to establish an auxiliary loss for training our model.
%
%By adopting the discovered inner structure as an auxiliary task, the obtained face representations are more discriminative \textit{w.r.t} the overall distribution in face space. Then our remaining challenge is how to robustly construct the graph based on unlabeled data.

Finding a reliable inner structure from noisy face data is non-trivial. % 
It is well-known that the representation induced by a single model is usually prone to bias and sensitive to noise.
To address the aforementioned challenge, we take a bottom-up approach to construct the graph by first identifying positive pairs reliably.
Specifically, we propose a novel \textbf{Consensus-Driven Propagation} (CDP)\footnote{Project page: http://mmlab.ie.cuhk.edu.hk/projects/CDP/}\footnote{Code: https://github.com/XiaohangZhan/cdp/} approach for graph construction in massive unlabeled data. It consists of two modules: a ``committee'' that provides multi-view information on the proposal pair, and a ``mediator'' that aggregates all the information for a final decision.

The ``\textbf{committee}'' module is inspired by query-by-committee (QBC)~\cite{seung1992query} that was originally proposed for active learning.
Different from QBC that measures disagreement, we collect consents from a committee, which comprises a base model and several auxiliary models. 
The heterogeneity of the committee reveals different views on the structure of the unlabeled data.
Then positive pairs are selected as the pair instances that the committee members most agree upon, rather than the base model is most confident of.
Hence the committee module is capable of selecting meaningful and hard positive pairs from the unlabeled data besides just easy pairs, complementing the model trained from just labeled data.
Beyond the simple voting scheme, as practiced by most QBC methods, we formulate a novel and more effective ``\textbf{mediator}'' to aggregate opinions from the committee.
The mediator is a binary classifier that produces the final decision as to select a pair or not.
We carefully design the inputs to the mediator so that it covers distributional information about the inner structure. The inputs include 1) voting results of the committee, 2) similarity between the pair, and 3) local density between the pair. The last two inputs are measured across all members of the committee and the base model.
%
%When comparing to directly voting, the ``mediator'' module is empirically proven to be more effective at selecting positive pairs \textit{w.r.t} both recall and precision.
%
Thanks to the ``committee'' module and the ``mediator'' module, we construct a robust consensus-driven graph on the unlabeled data.
Finally, we propagate pseudo-labels on the graph to form an auxiliary task for training our base model with unlabeled data.

To summarize, we investigate the usage of massive unlabeled data (over 6M images) for large-scale face recognition. 
Our setting closely resembles real-world scenarios where the unlabeled data are collected from unconstrained environments and their identities are exclusive from the labeled ones.  
We propose consensus-driven propagation (CDP) to tackle this challenging problem with two carefully-designed modules, the ``committee'' and the ``mediator'', which select positive face pairs robustly by aggregating multi-view information.
We show that a wise usage of unlabeled data can complement scarce manual labels to achieve compelling results.
With consensus-driven propagation, we can achieve comparable results by only using $9\%$ of the labels when compared to its fully-supervised counterpart.

\section{Related Work}

% %

\noindent\textbf{Semi-supervised Face Recognition.}
Semi-supervised learning~\cite{zhu2005semi,chapelle2009semi} is proposed to leverage large-scale unlabeled data, given a handful of labeled data.
It typically aims at propagating labels to the whole dataset from limited labels, by various ways, including self-training~\cite{yarowsky1995unsupervised,rosenberg2005semi}, co-training~\cite{blum1998combining,mitchell2004role}, multi-view learning~\cite{de1994learning}, expectation-maximization~\cite{dempster1977maximum} and graph-based methods~\cite{zhu2005semi-graph}.
For face recognition, Roli and Marcialis~\cite{roli2006semi} adopt a self-training strategy with PCA-based classifiers. In this work, the labels of unlabeled data are inferred with an initial classifier and are added to augment the labeled dataset.
Zhao \etal~\cite{zhao2011semi} employ Linear Discriminant Analysis (LDA) as the classifier and similarly use self-training to infer labels.
Gao \etal~\cite{gao2017semi} propose a semi-supervised sparse representation based method to handle the problem in few-shot learning that labeled examples are typically corrupted by nuisance variables such as bad lighting, wearing glasses.
All the aforementioned methods are based on the assumption that the set of categories are shared between labeled data and unlabeled data.
However, as mentioned before, this assumption is impractical when the quantity of face identities goes massive.

\noindent\textbf{Query-by-Committee}.
Query By Committee (QBC)~\cite{seung1992query} is a strategy relying on multiple discriminant models to explore disagreements, thus mining meaningful examples for machine learning tasks.
Argamon-Engelson \textit{et al.}~\cite{argamon1999committee} extend the QBC paradigm to the context of probabilistic classification and apply it to natural language processing tasks.
Loy \textit{et al.}~\cite{loy2012stream} extend QBC to discover unknown classes via a framework for joint exploration-exploitation active learning.
%
% Those previous works take advantages of the disagreements of the committee for threshold-free selection.
These previous works make use of the disagreements of the committee for threshold-free selection.
On the contrary, we exploit the consensus of the committee and extend it to the semi-supervised learning scenario.

\section{Methodology}
%
%To overcome the challenges of noisy unlabeled data and its unknown class space, we propose Consensus-Driven Propagation (CDP) to maximize the utility of massive unlabeled data for face recognition.

We first provide an overview of the proposed approach. 
Our approach consists of three stages:

\noindent
1) % training of base model and committee models on labeled data.
\textbf{Supervised initialization} - Given a small portion of labeled data, we separately train the base model and committee members in a fully-supervised manner.
More precisely, the base model $B$ and all the $N$ committee members $\{C_i \vert i=1,2,\dots,N\}$ learn a mapping from image space to feature space $\mathcal{Z}$ using labeled data $D_l$.
For the base model, this process can be denoted as the mapping: $\mathcal{F}_B: D_l \mapsto \mathcal{Z}$, and as for committee members: $\mathcal{F}_{C_i}: D_l \mapsto \mathcal{Z}$, $i=1,2,\dots,N$.

\begin{figure}[t]
\centering
\includegraphics[width=\linewidth]{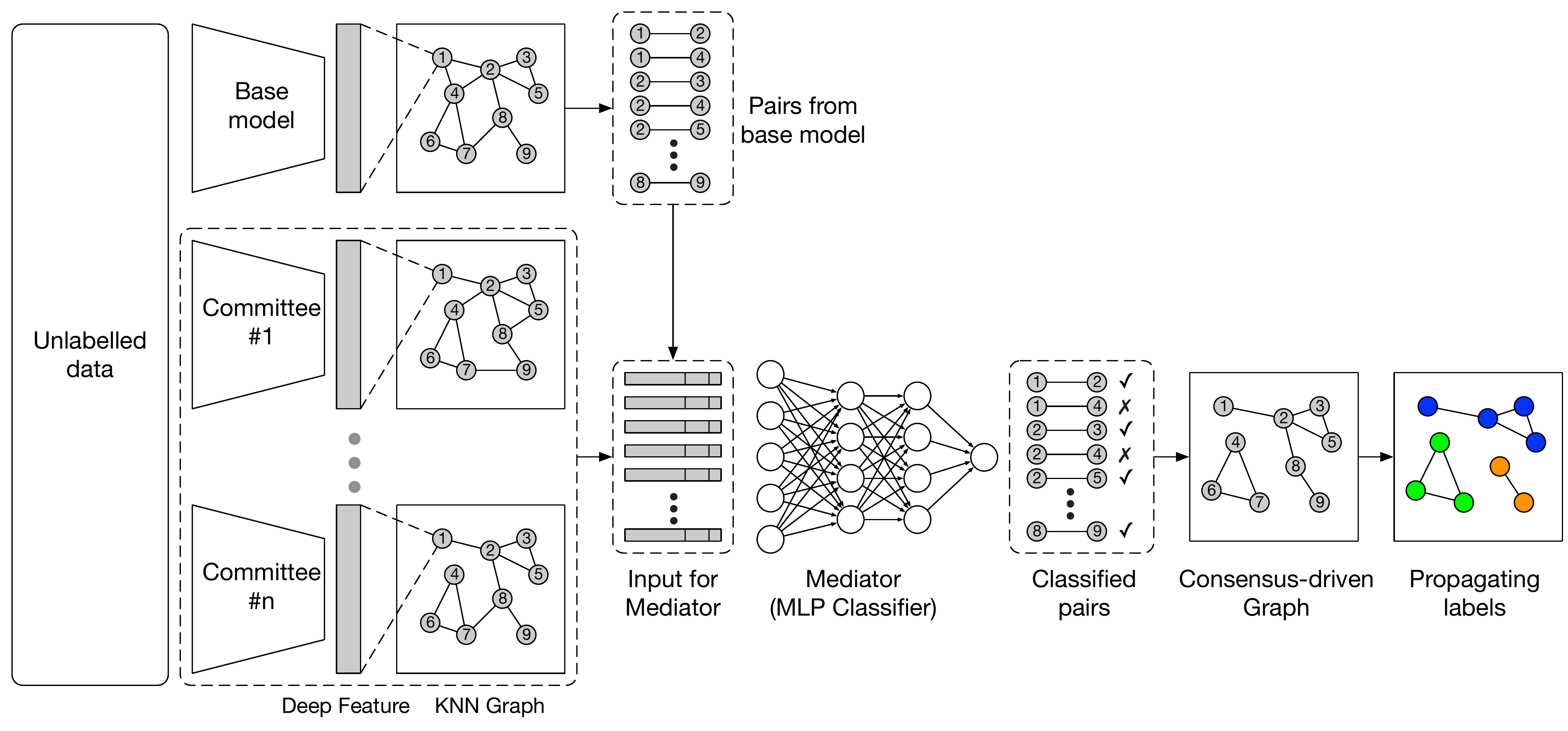}
\caption{\small{\textbf{Consensus-Driven Propagation}. We use a base model and committee models to extract features from unlabeled data and create k-NN graphs. 
%All pairs from base model k-NN graph are fed into an MLP classifier named ``mediator'' for binary classification. 
The input to the mediator is constructed by various local statistics of the k-NN graphs of the base model and committee. Pairs that are selected by the mediator compose the ``consensus-driven graph''. Finally, we propagate labels in the graph, and the propagation for each category ends by recursively eliminating low-confidence edges.}}
\label{fig:framework}
%\vspace{-0.5cm}
\end{figure}

\noindent
2) \textbf{Consensus-driven propagation} -
CDP is applied on unlabeled data to select valuable samples and conjecture labels thereon. The framework is shown in Fig.~\ref{fig:framework}.
We use the trained models from the first stage to extract deep features for unlabeled data and create k-NN graphs.
The ``committee'' ensures the diversity of the graphs.
Then a ``mediator'' network is designed to aggregate diverse opinions in the local structure of k-NN graphs to select meaningful pairs.
With the selected pairs, a consensus-driven graph is created on the unlabeled data and nodes are assigned with pseudo labels via our label propagation algorithm.

\noindent
3) \textbf{Joint training using labeled and unlabeled data} -
Finally, we re-train the base model with labeled data, and unlabeled data with pseudo labels, in a multi-task learning framework. 

\subsection{Consensus-Driven Propagation}
\label{sec:cdp}
In this section, we formally introduce the detailed steps of CDP.

%\vspace{0.1cm}
\noindent
\textbf{i. Building k-NN Graphs.}
For the base model and all committee members, we feed them with unlabeled data $D_u$ as input and extract deep features $\mathcal{F}_B\left (D_u\right )$ and $\mathcal{F}_{C_i}\left (D_u\right )$. With the features, we find $k$ nearest neighbors for each sample in $D_u$ by cosine similarity.
%
% suppose we use Cross-Entropy criterion when training the base model and committee members in the first stage.
%
This results in different versions of k-NN graphs, $\mathcal{G}_B$ for the base model and $\mathcal{G}_{C_i}$ for each committee member, totally $N+1$ graphs.
The nodes in the graphs are examples of the unlabeled data. Each edge in the k-NN graph defines a pair, and all the pairs from the base model's graph $\mathcal{G}_B$ form candidates for the subsequent selection, as shown in Fig.~\ref{fig:framework}.

%\vspace{0.1cm}
\noindent
\textbf{ii. Collecting Opinions from Committee.}
Committee members map the unlabeled data to the feature space via different mapping functions $\{\mathcal{F}_{C_i} \vert i=1,2,\dots,N$\}.
Assume two arbitrary connected nodes $n_0$ and $n_1$ in the graph created by the base model, and they are represented by different versions of deep features $\{\mathcal{F}_{C_i}(n_0) \vert i=1,2,\dots,N\}$ and $\{\mathcal{F}_{C_i}(n_1) \vert i=1,2,\dots,N\}$.
The committee provides the following factors:
%

%\vspace{0.1cm}
\noindent
1) The \textit{relationship}, $R$, between the two nodes.
Intuitively, it can be understood as whether two nodes are neighbors in the view of each committee member.
\begin{equation}
R_{C_i}^{\left(n_0, n_1\right)} =
    \begin{cases}
        1 & \text{if $\left(n_0, n_1\right) \in \mathcal{E}\left(\mathcal{G}_{c_{i}}\right) $} \\
        0 & \text{otherwise.}
    \end{cases},
    \quad i = 1,2,\dots,N,
\end{equation}
where $\mathcal{G}_{c_{i}}$ is the k-NN graph of $i$-th committee model and $\mathcal{E}$ denotes all edges of a graph.

%\vspace{0.1cm}
\noindent
2) The \textit{affinity}, $A$, between the two nodes.
It can be computed as the similarity measured in the feature space with the mapping functions defined by the committee members. Assume that we use cosine similarity as a metric,
\begin{equation}
A_{C_i}^{\left(n_0, n_1\right)} = \cos\left(\left\langle \mathcal{F}_{C_i}\left(n_0\right), \mathcal{F}_{C_i}\left(n_1\right)\right\rangle\right), \quad i = 1,2,\dots,N. \\
\end{equation}

%\vspace{0.1cm}
\noindent
3) The \textit{local structures} \textit{w.r.t} each node. This notion can refer to the distribution of a node's first-order, second-order, and even higher-order neighbors.
Among them the first-order neighbors play the most important role to represent the ``local structures'' \textit{w.r.t} a node.
And such distribution can be approximated as the distribution of similarities between the node $x$ and all of its neighbors $x_k$, where $k=1,2,...,K$.
\begin{equation}
D_{C_i}^x = \{\cos\left(\left\langle \mathcal{F}_{C_i}\left(x\right), \mathcal{F}_{C_i}\left(x_k\right)\right\rangle\right), k = 1,2,\dots,K\}, \quad i=1,2,\dots,N.
\end{equation}

\begin{figure}[t]
\begin{center}
%\vspace{8cm}
\includegraphics[width=\linewidth]{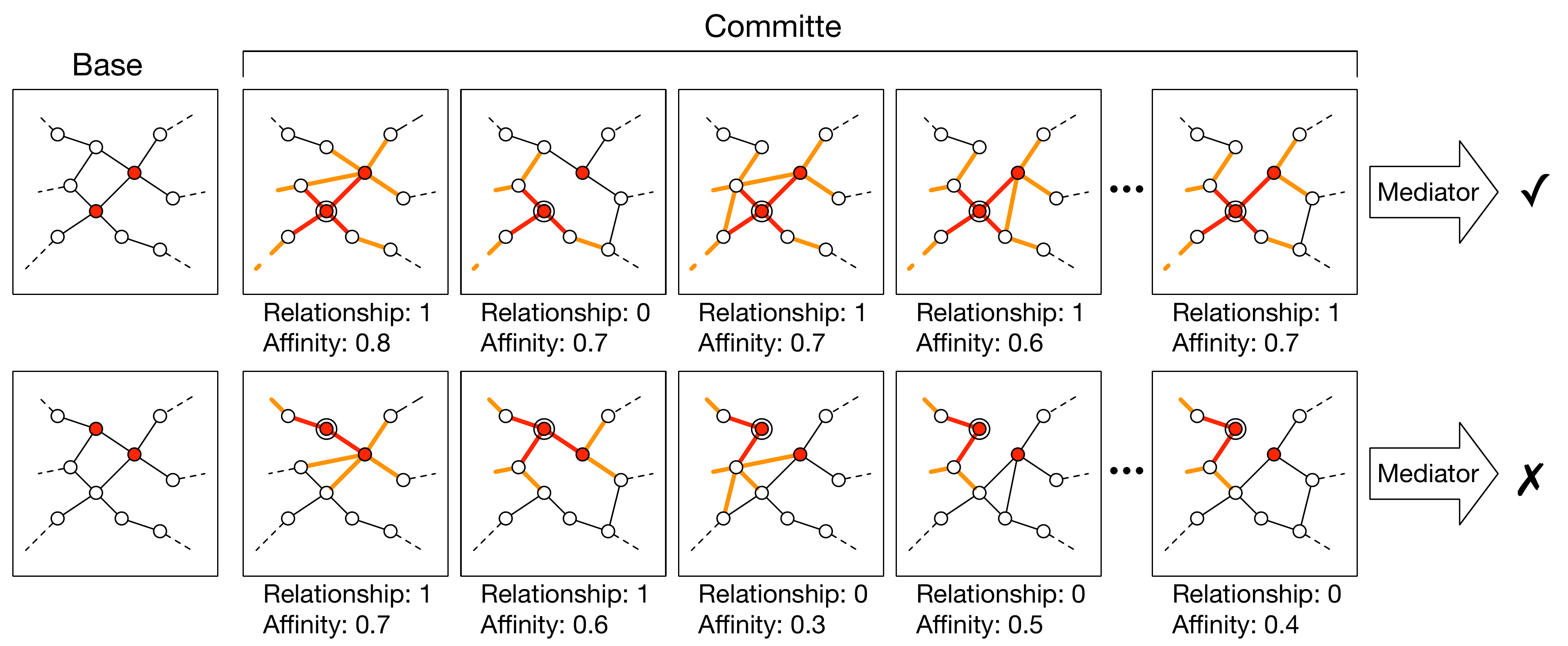}
%\vskip -0.2cm
\caption{\textbf{Committee and Mediator}. This figure illustrates the mechanisms of committee and mediator. The figure shows some sampled nodes in different versions of graphs brought by the base model and the committee. In each row, the two red nodes are candidate pairs. The pair in the first row is classified as positive by the mediator, while the pair in the second row is considered as negative. The committee provides diverse opinions on ``relationship'', ``affinity'', and ``local structure''. The ``local structure'' is represented as the distribution of first-order (red edges) and second-order (orange edges) neighbors. Note that the figure only shows the ``local structure'' centered on one of the two nodes (the node with double circles).}
\label{fig:mediator}
\end{center}
%\vspace{-0.5cm}
\end{figure}

As illustrated in Fig.~\ref{fig:mediator},  given a pair of nodes extracted from the base model's graph, the committee members provide diverse opinions to the \textit{relationships}, the \textit{affinity} and the \textit{local structures}, due to their nature of heterogeneity.
%
%What we care about is the consensus part of these opinions which is of the essence for pair selection.
%
%Then how to appropriately aggregate these opinions?
%
From these diverse opinions, we seek to find a consent through a mediator in the next step.

%\vspace{0.1cm}
\noindent
\textbf{iii. Aggregate Opinions via Mediator.}
The role of a mediator is to aggregate and convey committee members' opinions for pair selection.
We formulate the mediator as a Multi-Layer Perceptron (MLP) classifier albeit other types of classifier are applicable.
Recall that all pairs extracted from the base model's graph constitute the candidates.
The mediator shall re-weight the opinions of the committee members and make a final decision by assigning a probability to each pair to indicate if a pair shares the same identity, \textit{i.e.}, positive, or have different identities, \textit{i.e.}, negative.

The input to the mediator for each pair $\left(n_0, n_1\right)$ is a concatenated vector containing three parts (here we denote $B$ as $C_0$ for simplicity of notation):

\noindent
1) ``relationship vector'' $I_R\in \mathbb{R}^{N}$: $I_R = \left(\dots R_{C_i}^{\left(n_0, n_1\right)}\dots\right), i=1,2,\dots,N$, from the committee.

\noindent
2) ``affinity vector'' $I_A\in \mathbb{R}^{N+1}$: 
%$I_A = \left(A_B^{\left(n_0, n_1\right)},\dots A_{C_i}^{\left(n_0, n_1\right)}\dots \right), i=1,2,\dots,N,$ 
$I_A = \left(\dots A_{C_i}^{\left(n_0, n_1\right)}\dots \right), i=0,1,2,\dots,N,$ 
from both the base model and the committee.

\noindent
3) ``neighbors distribution vector'' including ``mean vector'' $I_{D_{mean}}\in \mathbb{R}^{2\left(N+1\right)}$ and ``variance vector'' $I_{D_{var}}\in \mathbb{R}^{2\left(N+1\right)}$:
\begin{equation}
\begin{split}
%& I_{D_{mean}} = \left(E\left(D_B^{n_0}\right), E\left(D_B^{n_1}\right), \dots %E\left(D_{C_i}^{n_0}\right)\dots, \dots %E\left(D_{C_i}^{n_1}\right)\dots\right), i=1,2,\dots,N, \\
%& I_{D_{var}} = \left(\sigma\left(D_B^{n_0}\right), %\sigma\left(D_B^{n_1}\right), \dots \sigma\left(D_{C_i}^{n_0}\right)\dots, %\dots \sigma\left(D_{C_i}^{n_1}\right)\dots\right), i=1,2,\dots,N,
& I_{D_{mean}} = \left(\dots E\left(D_{C_i}^{n_0}\right)\dots ~,~ \dots E\left(D_{C_i}^{n_1}\right)\dots\right), i=0,1,2,\dots,N, \\
& I_{D_{var}} = \left(\dots \sigma\left(D_{C_i}^{n_0}\right)\dots ~,~ \dots \sigma\left(D_{C_i}^{n_1}\right)\dots\right), i=0,1,2,\dots,N,
\end{split}
\end{equation}
from both the base model and the committee for each node.
Then it results in $6N+5$ dimensions of the input vector.
The mediator is trained on $D_l$, and the objective is to minimize the corresponding Cross-Entropy loss function.
For testing, pairs from $D_u$ are fed into the mediator and those with a high probability to be positive are collected.
Since most of the positive pairs are redundant, we set a high threshold to select pairs, thus sacrificing recall to obtain positive pairs with high precision.

%\vspace{0.1cm}
\noindent
\textbf{iv. Pseudo Label Propagation.}
The pairs selected by the mediator in the previous step compose a ``Consensus-Driven Graph'', whose edges are weighted by pairs' probability to be positive.
Note that the graph does not need to be a connected graph. Unlike conventional label propagation algorithms, we do not assume labeled nodes on the graph. To prepare for subsequent model training, we propagate pseudo labels based on the connectivity of nodes.
To propagate pseudo labels, we devise a simple yet effective algorithm to identify connected components.
At first, we find connected components based on the current edges in the graph and add it to a queue.
For each identified component, if its node number is larger than a pre-defined value, we eliminate low-score edges in the component, find connected components from it, and add the new disjoint components to the queue.
If the node number of a component is below the pre-defined value, we annotate all nodes in the component with a new pseudo label.
We iterate this process until the queue is empty when all the eligible components are labeled.

\subsection{Joint Training using Labeled and Unlabeled Data}
\begin{figure}[t]
\centering
\includegraphics[width=\linewidth]{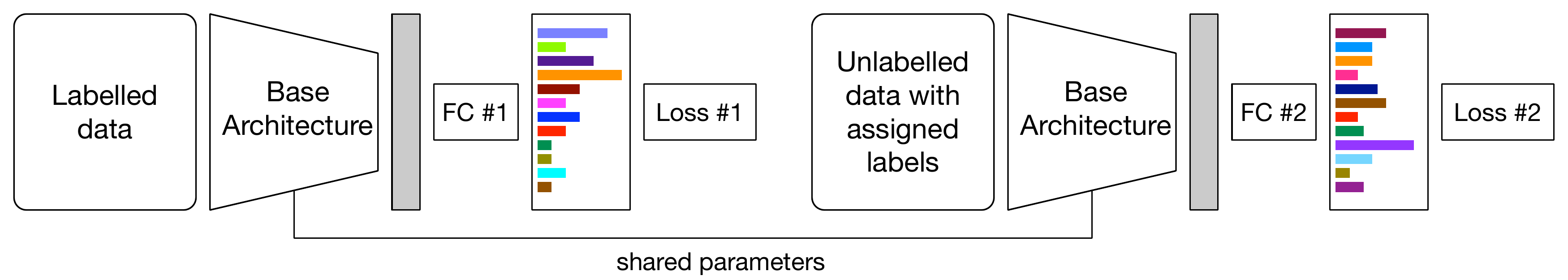}
\caption{\small{Model updating in multi-task fashion. The weights of two CNNs are shared. ``FC'' denotes fully-connected classifier. In our experiments we use weighted Cross-Entropy loss as the objective.}}
\label{fig:multitask}
%\vspace{-0.5cm}
\end{figure}

Once the unlabeled data are assigned with pseudo labels, we can use them to augment the labeled data and update the base model.
Since the identity intersection of two data sets is unknown, we formulate the learning in a multi-task training fashion, as shown in Fig.~\ref{fig:multitask}.
The CNN architectures for the two tasks are exactly the same as the base model, and the weights are shared.
Both CNNs are followed by a fully-connected layer to map deep features into the respective label space.
The overall optimization objective is $\mathcal{L} = \lambda \sum\nolimits_{x_l,y_l} \ell\left(x_l, y_l\right) + \left(1-\lambda\right) \sum\nolimits_{x_u,y_a} \ell\left(x_u, y_a\right)$,
%\begin{equation}
%\label{eqn:loss}
%\mathcal{L} = \lambda \sum\nolimits_{x_l,y_l} \ell\left(x_l, y_l\right) + \left(1-\lambda\right) \sum\nolimits_{x_u,y_a} \ell\left(x_u, y_a\right),
%\end{equation}
where the loss, $\ell(\cdot)$, is the same as the one for training the base model and committee members. 
In the following experiments, we employ $softmax$ as our loss function. But note that there is no restriction to which loss is equipped with CDP. In Section~\ref{sec:analysis}, we show that CDP still helps considerably despite with advanced loss functions.
In this equation, $\{x_l, y_l\}$ denotes labeled data, while $\{x_u, y_a\}$ denotes unlabeled data and the assigned labels.
$\lambda \in \left(0,1\right)$ is the weight to balance the two components. Its value is fixed following the proportion of images in the labeled and unlabeled set.
The model is trained from scratch.

\section{Experiments}

\label{sec:expsetup}

\noindent
\textbf{Training Set.}
MS-Celeb-1M~\cite{guo2016ms} is a large-scale face recognition dataset containing $10M$ training examples with $100K$ identities.
To address the original annotation noises, we clean up the official training set and crawl images of more identities, producing about $7M$ images with $385K$ identities.
We split the cleaned dataset into 11 balanced parts randomly by identities, so as to ensures that there is no identity overlapping between different parts.
Note that though our experiments adopt this harder setting, our approach can be readily applied to identity-overlapping settings since it makes no assumptions on the identities.
Among the different parts, one part is regarded as labeled and the other ten parts are regarded as unlabeled.
We also use one of the unlabeled parts as a validation set to adjust hyper-parameters and perform ablation study.
The labeled part contains $634K$ images with $35,012$ identities. 
%
% It's relatively small among modern face recognition works.
%
The model trained only on the labeled part is regarded as the lower bound performance. The fully-supervised version is trained with full labels from all the 11 parts.
To investigate the utility of the unlabeled data, we compare different methods with 2, 4, 6, 8, and 10 parts of unlabeled data included, respectively.

\noindent
\textbf{Testing Sets.}
MegaFace~\cite{kemelmacher2016megaface} is currently the largest public benchmark for face identification.
It includes a gallery set containing $1M$ images, and a probe set from FaceScrub~\cite{ng2014data} with 3,530 images.
However, there are some noisy images from FaceScrub, hence we use the noises list proposed by InsightFace\footnote{InsightFace: https://github.com/deepinsight/insightface/tree/master/src/megaface} to clean it.
We adopt rank-1 identification rate in MegaFace benchmark, which is to select the top-1 image from the $1M$ gallery and average the top-1 hit rate.
IJB-A~\cite{ng2014data} is a face verification benchmark contains 5,712 images from 500 identities.
%
% Face verification is to determine if two face images are belong to the same person.
%
We report the \textit{true positive rate} under the condition that the \textit{false positive rate} is 0.001 for evaluation.

\noindent
\textbf{Committee Setup.}
To create a ``committee'' with high heterogeneity, we employ popular CNN architectures including \textit{ResNet18}~\cite{he2016deep}, \textit{ResNet34}, \textit{ResNet50}, \textit{ResNet101}, \textit{DenseNet121}~\cite{iandola2014densenet}, \textit{VGG16}~\cite{simonyan2014very}, \textit{Inception V3}~\cite{szegedy2016rethinking}, \textit{Inception-ResNet V2}~\cite{szegedy2017inception} and a smaller variant of \textit{NASNet-A}~\cite{zoph2017learning}.
The number of committee members is eight in our experiments, but we also explore the choice of the number of committee member from $0$ to $8$.
We trained all the architectures with the labeled part of data and the performance is listed in Table~\ref{tab:committee}. The numbers of parameters are also listed.
\textit{Tiny NASNet-A} shows the best performance among all the architectures but uses the smallest number of parameters.
Model ensemble results are also presented. Empirically, the best ensemble combination is to assemble the four top-performing models, i.e., \textit{Tiny NASNet-A}, \textit{Inception-Resnet V2}, \textit{DenseNet121}, \textit{ResNet101}, yielding 68.86\% and 76.97\% on two benchmarks.
We select \textit{Tiny NASNet-A} as our base architecture and the other 8 models as committee members.
The following experiments demonstrate that the ``committee'' helps even though its members are weaker than the base architecture.
%
% In Section~\ref{sec:analysis} we also switch the base model and compare the performances.
In Section~\ref{sec:analysis} we also show that our approach is widely applicable by switching the base architecture.

\begin{table}[tb]
	\centering
	\caption{Performance and the number of parameters of the base model and the committee members.}
	\label{tab:committee}
\scriptsize{ 	
	\begin{tabular}{l|c|c|c|c}
		\hline
									& Architecture 		& MegaFace	& IJB-A	& Parameters 	\\ \hline
		Base						& Tiny NASNet-A 	& \textbf{61.78} 	& \textbf{75.87} & $20.1M$		\\ \hline \hline
		\multirow{8}{*}{Committee}	& VGG16				& 50.22		& 70.75	& $75.6M$		\\ \cline{2-5}
									& ResNet18			& 51.48		& 69.23	& $23.5M$		\\ \cline{2-5}
									& ResNet34			& 52.44		& 72.52	& $33.6M$		\\ \cline{2-5}
									& Inception V3		& 52.82		& 75.53	& $33.0M$		\\ \cline{2-5}
									& ResNet50			& 56.16		& 73.21	& $36.3M$		\\ \cline{2-5}
									& ResNet101			& 57.87		& 74.52	& $55.3M$		\\ \cline{2-5}
									& Inception-ResNet V2 & 58.68	& 75.13	& $66.1M$		\\ \cline{2-5}
									& DesNet121			& 60.77		& 69.78	& $28.9M$		\\ \hline\hline
		Ensemble					& (multiple)		& 69.86		& 76.97	&	-			\\ \hline 
	\end{tabular}
}
\vspace{-0.5cm}
\end{table}

\noindent\textbf{Implementation Details.}
The ``mediator'' is an MLP classifier with $2$ hidden layers, each of which containing $50$ nodes.
It uses ReLU as the activation function.
At test time, we set the probability threshold as $0.96$ to select high-confident pairs. More details can be found in the supplementary material.

\subsection{Comparisons and Results}

\noindent
\textbf{Competing Methods.}
1) \textit{Supervised deep feature extractor + Hierarchical Clustering}:
We prepare a strong baseline by hierarchical clustering with supervised deep feature extractor.
Hierarchical clustering is a practical way to deal with massive data comparing to other clustering methods.
The clusters are assigned pseudo labels and augment the training set.
For best performance, we carefully adjust the threshold of hierarchical clustering using the validation set and discard clusters with just a single image.
2) \textit{Pair selection by naive committee voting}:
A pair is selected if this pair is voted by all the committee members (best setting empirically). A vote is counted if there is an edge in the k-NN graph of a committee member.

\begin{figure}[t]
\centering
\includegraphics[width=\linewidth]{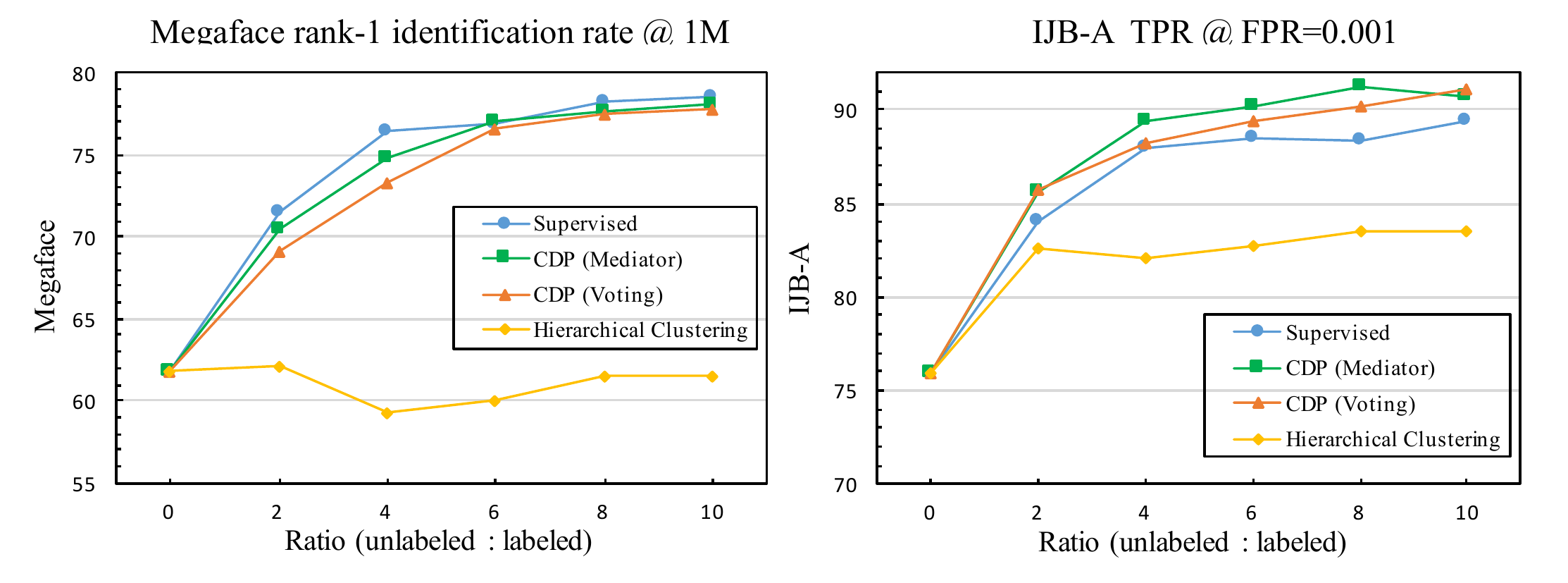}
%\vspace{-0.5cm}
\caption{\small{Performance comparison on MegaFace identification task and IJB-A verification task with different ratios of unlabeled data added to one portion of labeled data. CDP is proven to 1) obtain large improvements over the lower bound (ratio of unlabeled:labeled is 0:1); 2) surpass the clustering method by a large margin; 3) obtain competitive or even higher results over the fully-supervised counterpart.}}
\label{fig:curves}
%\vspace{-0.5cm}
\end{figure}

\noindent
\textbf{Benchmarking.}
As shown in Fig.~\ref{fig:curves}, the proposed CDP method achieves impressive results on both benchmarks.
From the results, we observe that:

\noindent
1) Comparing to the lower bound (ratio of unlabeled:labeled is 0:1) with no unlabeled data, CDP obtains significant and steady improvements given different quantities of unlabeled data.

\noindent
2) CDP surpasses the baseline ``Hierarchical Clustering'' by a large margin, obtaining competitive or even better results over the fully-supervised counterpart. In the MegaFace benchmark, with 10 fold unlabeled data added, CDP yields 78.18\% of identification rate. Comparing to the lower bound without unlabeled data that yields 61.78\%, CDP obtains 16.4\% of improvement. Notably, there are only 0.34\% gap between CDP and the fully-supervised setting that reaches 78.52\%. The results suggest that CDP is capable of maximizing the utility of the unlabeled data.

\noindent
3) CDP by the ``mediator'' performs better than by naive voting, indicating that the ``mediator'' is more capable in aggregating committee opinions.

\noindent
4) In the IJB-A face verification task, both settings of CDP surpass the fully-supervised counterpart. The poorer results observed on the fully-supervised baseline suggest the vulnerability of this task against noisy annotations in the training set, as discussed in Section~\ref{sec:introduction}. By contrast, our method is more resilient to noise. We will discuss this next based on Fig.~\ref{fig:vlz_denoise}.

\noindent
\textbf{Visual Results.}
We visualize the results of CDP in Fig.~\ref{fig:vlz_denoise}.
It can be observed that CDP is highly precise in identity label assignment, regardless the diverse backgrounds, expressions, poses and illuminations.
It is also observed that CDP behaves to be selective in choosing samples for pair candidates, as it automatically discards 1) wrongly-annotated faces not belonging to any identity; 2) samples with extremely low quality, including heavily blurred and cartoon images.
This explains why CDP outperforms the fully-supervised baseline in the IJB-A face verification task (Fig.~\ref{fig:curves}).
%Hence CDP is a convenient way to filter meaningful unlabeled data.

%

\subsection{Ablation Study}
\label{sec:ablation}
We perform ablation study on the validation set to show the gain of each component, as shown in ~\ref{tab:ablation}.
Several indicators are included for comparison.
Higher recall and precision of selected pairs will result in better consensus-driven graph, hence improves the quality of assigned labels.
For assigned labels, pairwise recall and precision reflect the quality of the labels, and directly correlate the final performance on two benchmarks.
Higher pairwise recall indicates more true examples in a category, which is important for the subsequent training.
Higher pairwise precision indicates less noises in a category.

\noindent
\textbf{The Effectiveness of ``Committee''.}
When we vary the number of committee members, we adjust pair similarity threshold to obtain fixed recall for convenience.
With increasing committee number, an interesting observation is that, the peak of precision occurs where the number is 4.
However, it does not bring the best quality of assigned labels, which occurs where the number is 6-8.
This shows that more committee members will bring more meaningful pairs rather than just correct pairs.
This conclusion is consistent with our assumption that the committee is able to select more hard positive pairs relative to the base model.

\noindent
\textbf{The Effectiveness of ``Mediator''.}
For the ``mediator'', we study the influence of different input settings.
With only the ``relationship vector'' $I_R$ as input, the values of those indicators are close to that of direct voting.
Then the ``affinity vector'' $I_A$ remarkably improves recall and precision of selected pairs, and also improves both pairwise recall and precision of assigned labels.
The ``neighbors distribution vector'' $I_{D_{mean}}$ and $I_{D_{var}}$ further boost the quality of the assigned labels.
The improvements originate in the effect brought by these aspects of information, and hence the ``mediator'' performs better than naive voting.

\begin{table}[tb]
\centering
\caption{Ablation study on validation set. $I_R$: ``relationship vector'', $I_A$: ``affinity vector'', $I_D$: ``neighbors distribution vector''. Among the indicators pairwise recall and precision for assigned labels directly correlate the benchmarking results. It is concluded that more committee members bring more meaningful pairs rather than just correct pairs, and the ``mediator'' is capable in aggregating multiple aspects of consensus information.}
\label{tab:ablation}
\scriptsize{
\begin{tabular}{l|c|c|c|c|c|c|c}
\hline
\multirow{3}{*}{Methods}   & \multirow{3}{*}{\begin{tabular}[c]{@{}c@{}}Committee\\ number\end{tabular}}   & \multirow{3}{*}{\begin{tabular}[c]{@{}c@{}}Mediator\\ inputs\end{tabular} } & \multicolumn{3}{c|}{Pair selection} 	& \multicolumn{2}{c}{Assigned labels}   \\ \cline{4-8}
						  &						&				& \begin{tabular}[c]{@{}c@{}}pair\\ number\end{tabular} & recall & precision & \begin{tabular}[c]{@{}c@{}}pairwise\\ recall\end{tabular}  & \begin{tabular}[c]{@{}c@{}}pairwise\\ precision\end{tabular} 	\\ \hline
Clustering              & -                  & -              & - & -       &  -   	&  0.558		& 0.950      	\\ \hline\hline
\multirow{5}{*}{Voting}     & 0                  & -           & $1.4M$ & 0.313   &  0.966	& 0.680 & 0.829      	\\ \cline{2-8} 
						  & 2                  & -              &$1.4M$ & 0.313   &  0.986	& 0.783 & 0.849      	\\ \cline{2-8} 
                          & 4                  & -             &$1.4M$ & 0.313   &  \textbf{0.987}	& 0.791 & 0.862		   	\\ \cline{2-8} 
                          & 6                  & -             &$1.4M$ & 0.313   &  0.984  & 0.801	& \textbf{0.877}  		\\ \cline{2-8} 
                          & 8                  & -             & $1.4M$& 0.313   &  0.979  & \textbf{0.807} & 0.876 		\\ \hline\hline
\multirow{3}{*}{Mediator} & \multirow{3}{*}{8} & $I_R$            & $1.4M$ & 0.318   &  0.975	& 0.825 & 0.822 		\\ \cline{3-8} 
                          &                    & $I_R$+$I_A$      & $2.5M$ & \textbf{0.561}   & 0.982   & \textbf{0.832} & 0.888 		\\ \cline{3-8} 
                          &                    & $I_R$+$I_A$+$I_D$ & $2.4M$& 0.527   & \textbf{0.983}	& 0.825 & \textbf{0.912}  		\\ \hline
\end{tabular}
}
\end{table}

\subsection{Further Analysis}
\label{sec:analysis}

\noindent\textbf{Different Base Architectures.}
In previous experiments we have chosen \textit{Tiny NASNet-A} as the base model and other architecture as committee members.
%
% Different architectures have various capacities.
%
To investigate the influence of the base model, here we switch the base model to \textit{ResNet18}, \textit{ResNet50}, \textit{Inception-ResNet V2} respectively and list their performance in Table~\ref{tab:varybase}.
We observe consistent and large improvements from the lower bound on all the base architectures.
Specifically, with high-capacity \textit{Inception-ResNet V2}, our CDP achieves 81.88\% and 92.07\% on MegaFace and IJB-A benchmarks, with 23.20\% and 16.94\% improvements.
It is significant considering that CDP uses the same amount of labeled data as the lower bound ($9\%$ of all the labels).
Our performance is also much higher than the ensemble of base model and committee, indicating that CDP actually exploits the intrinsic structure of the unlabeled data to learn effective representations.

\begin{table}[tb]
\centering
\caption{The comparison of different base architectures. Lower bound: the models trained on 1-fold labeled data only; CDP: our semi-supervised models with 1-fold labeled data and 10-fold unlabeled data; Supervised: the models trained on all the 11-fold data with labels. With higher-capacity architectures, CDP achieves even larger improvements.}
\label{tab:varybase}
\scriptsize{
\begin{tabular}{l|c|c|c|c|c|c|c|c}
\hline
\multirow{2}{*}{Base}        & \multicolumn{2}{c|}{ResNet18} & \multicolumn{2}{c|}{ResNet50} & \multicolumn{2}{c|}{Tiny NASNet-A} & \multicolumn{2}{c}{Inception-ResNet V2} \\ \cline{2-9}
            & MegaFace        & IJB-A       & MegaFace        & IJB-A       & MegaFace          & IJB-A          & MegaFace             & IJB-A             \\ \hline
Lower Bound & 51.48           & 69.23       & 56.16           & 73.12       & 61.78             & 75.87          & 58.68                & 75.13             \\ \hline
CDP    		& 72.75           & 86.23       & 75.66           & 88.34       & 78.18             & 90.64          & 81.88                & 92.07             \\ \hline
Supervised  & 73.88           & 85.08       & 77.13           & 87.92       & 78.52             & 89.40          & 84.74                & 91.90             \\ \hline
\end{tabular}
}
\end{table}

\noindent\textbf{Different $k$ in k-NN.}
Here we inspect the effect of $k$ in k-NN. In this comparable study, the probability threshold of a pair to be positive is fixed to $0.96$.
As shown in Table~\ref{tab:k}, higher $k$ results in more selected pairs and thus a denser consensus-driven graph, but the precision is almost unchanged.
Note that the recall drops because the cardinal true pair number increases faster than the that of selected pairs.
Actually, it is unnecessary to pursue high recall rate if the selected pairs are enough.
For assigned labels, denser graph brings higher pairwise recall and lower precision.
Hence it is a trade-off between pairwise recall and precision of the assigned labels via varying $k$.
\begin{table}[tb]
\begin{minipage}[c]{.6\linewidth}%
	\centering
	\caption{The influence of $k$ in k-NN. Varying $k$ provides a trade-off between pairwise recall and precision of the assined labels.}
	\label{tab:k}
	\scriptsize{
	\begin{tabular}{c|c|c|c|c|c}
	\hline
	\multirow{2}{*}{$k$} & \multicolumn{3}{c|}{Pair selection} & \multicolumn{2}{c}{Assigned labels} \\ \cline{2-6} 
	                   & \begin{tabular}[c]{@{}c@{}}pair\\ number\end{tabular} & recall 	& precision & \begin{tabular}[c]{@{}c@{}}pairwise\\ recall\end{tabular} & \begin{tabular}[c]{@{}c@{}}pairwise\\ precision\end{tabular} 	\\ \hline
	10                 & 1.61M		& 	\textbf{0.601}   &  \textbf{0.985}    &  0.810   	& \textbf{0.940}    	\\ \hline
	20                 & 2.54M		&   0.527 	&  0.983    &  0.825 	& 0.912    	\\ \hline
	30                 & 2.96M		&   0.507  	&  0.982    &  0.834  	& 0.886    	\\ \hline
	40                 & \textbf{3.17M}		&   0.464  	&  0.982  	&  \textbf{0.837}  	& 0.874  	\\ \hline
	\end{tabular}
	}
\end{minipage}
\begin{minipage}[c]{.35\linewidth}%
	\centering
	%\vspace{0.4cm}
	\includegraphics[width=0.8\linewidth]{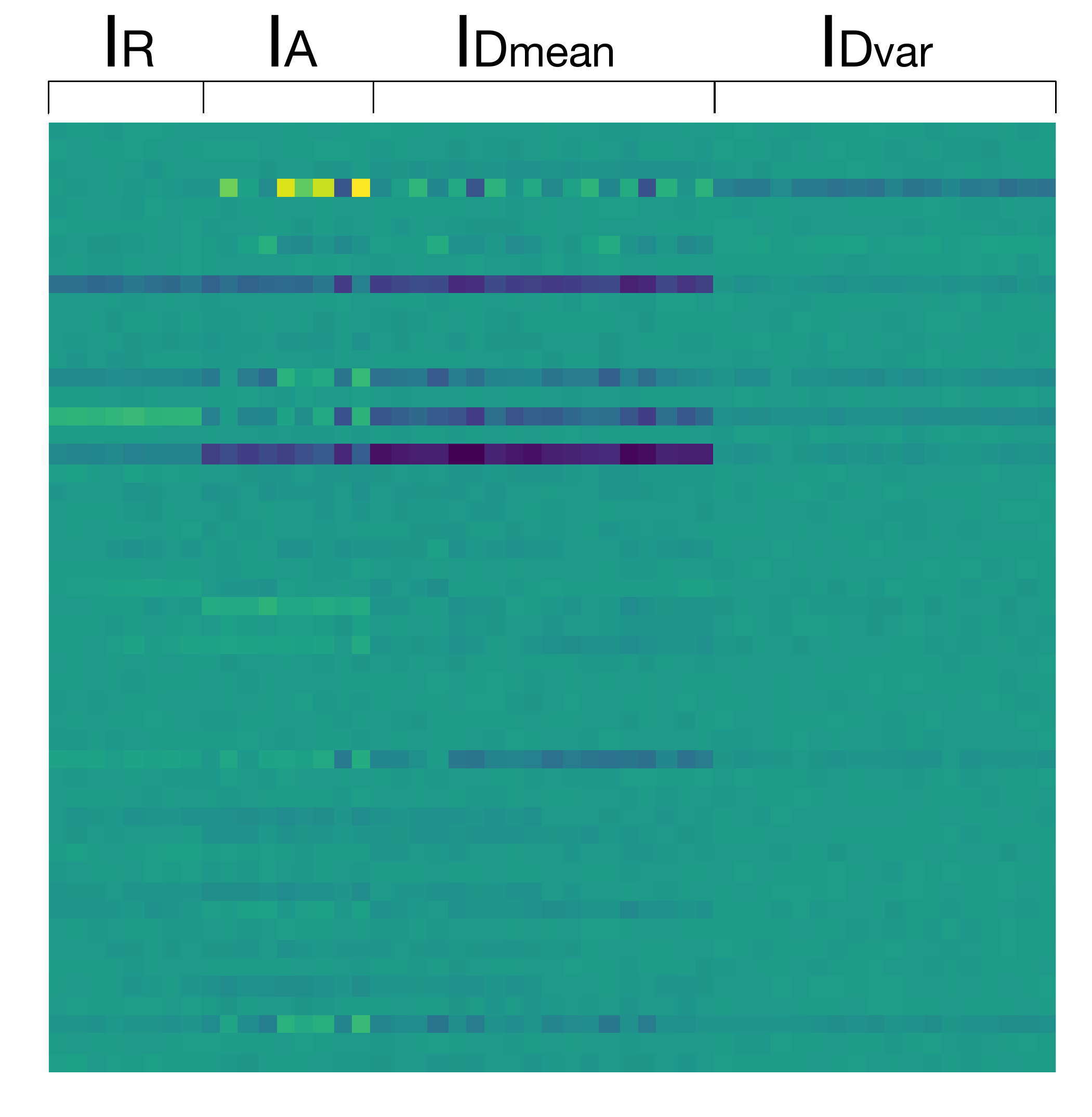}
	\captionof{figure}{\small{Mediator Weights.}}
	\label{fig:vlz_weight}
\end{minipage}
\end{table}
	
\noindent\textbf{Committee Heterogeneity.}
To study the influence of committee heterogeneity, we conduct experiments with homogeneous committee architectures.
The homogeneous committee consists of eight \textit{ResNet50} models that are trained with different data feeding orders, and the base model is the identical one as the heterogeneous setting.
The model capacity of \textit{ResNet50} is at the median of the heterogeneous committee, for a fair comparison.
As shown in Table~\ref{tab:heterogeneity}, heterogeneous committee performs better than the homogeneous one via either voting or the ``mediator''.
The study verifies that committee heterogeneity is helpful.

\begin{table}[tb]
\centering
\caption{The influence of committee heterogeneity. As a comparison, the heterogeneous committee performs better than the homogeneous committee.}
\label{tab:heterogeneity}
\scriptsize{
\begin{tabular}{l|l|c|c|c|c|c}
\hline
\multirow{3}{*}{Committee} & \multirow{3}{*}{Methods}	 & \multicolumn{3}{c|}{Pair selection} 	& \multicolumn{2}{c}{Assigned labels}   \\ \cline{3-7}
			 & & pair number & recall & precision & \begin{tabular}[c]{@{}c@{}}pairwise\\ recall\end{tabular}  & \begin{tabular}[c]{@{}c@{}}pairwise\\ precision\end{tabular} 	\\ \hline
\multirow{2}{*}{Homogeneous}    & voting 	& 1.93M & 0.368 	& 0.648   	& 0.746  & 0.681   		\\ \cline{2-7}
								& mediator 	& 2.46M & 0.508 	& 0.853   	& 0.798  & 0.831   		\\ \hline 
\multirow{2}{*}{Heterogeneous}  & voting 	&	1.41M & 0.313   &  0.979  	& 0.807 & 0.876 		\\ \cline{2-7}
								& mediator 	&	2.54M & \textbf{0.527}  & \textbf{0.983}		& \textbf{0.825}   & \textbf{0.912} 		\\ \hline 
\end{tabular}
}
\end{table}

\noindent\textbf{Inside Mediator.}
To evaluate the participation of each input, we visualize the first layer's weights in the ``mediator'', as shown in Fig.~\ref{fig:vlz_weight}.
It is the $50\times53$ weights of the first layer in the ``mediator'', where the number of input and output channels is 53 and 50. Hence each column represents the weights of each input. The values in green is close to 0, and blue less than 0, yellow greater than 0. Both values in yellow and blue indicate high response to the corresponding inputs. %(Best viewed in color)
We conclude that the committee's ``affinity vector'' ($I_A$) and the mean vector of ``neighbors distribution'' ($I_{D_{mean}}$) contribute higher to the response, than ``relationship vector'' ($I_R$) and the variance vector of ``neighbors distribution'' ($I_{D_{var}}$).
The result is reasonable since similarities contain more information than voting results, and the mean of neighbors' distribution directly reflects the local density.

\noindent\textbf{Incorporating Advanced Loss Functions.}
Our CDP framework is compatible with various forms of loss functions.
Apart from $softmax$, we also equip CDP with an advanced loss function, ArcFace~\cite{deng2018arcface}, the current top entry on MegaFace benchmark.
For parameters related to ArcFace, we set the margin $m=0.5$ and adopt the output setting ``E'', that is ``BN-Dropout-FC-BN''.
We also use a cleaner training set aiming to obtain a higher baseline.
As shown in Table ~\ref{tab:arcface}, we observe that CDP still brings large improvements over this much higher baseline.
%
%\vspace{-4pt}
\begin{table}[h!]
	\centering
	\vspace{-4pt}
	\caption{Comparisons of the gain brought by CDP with 2-folds unlabeled data between the previous baseline (Softmax) and the new baseline (ArcFace~\cite{deng2018arcface} with a cleaner training set). The performances are reported on MegaFace test set.}
	\label{tab:arcface}
	\scriptsize{
	\begin{tabular}{c|c|c}
		\hline
		& ~Softmax~			& ~ArcFace~\cite{deng2018arcface}~  \\\hline
		~baseline~   		& 61.78\%  	& 76.93\% \\\hline
		~CDP (~Ratio~=~2)~ 	& 70.51\%   & 83.68\% \\\hline
	\end{tabular}
	}
	\vspace{-4pt}
\end{table}

\noindent\textbf{Efficiency and Scalability.}
The step-by-step runtime of CDP is listed as follows: for million-level data, graph construction (k-NN search) takes $4$ minutes to perform on a CPU with $48$ processors, the ``committee''+``mediator'' network inference takes $2$ minutes to perform on eight GPUs, and the propagation takes another $2$ minutes on a single CPU. %This is much faster than traditional clustering-based methods, which usually take several hours to complete. 
Since our approach constructs graphs in a bottom-up manner and the ``committee''+``mediator'' only operate on local structures, the runtime of CDP grows linearly with the number of unlabeled data. Therefore, CDP is both efficient and scalable.

\begin{figure}[t]
\centering
\includegraphics[width=\linewidth]{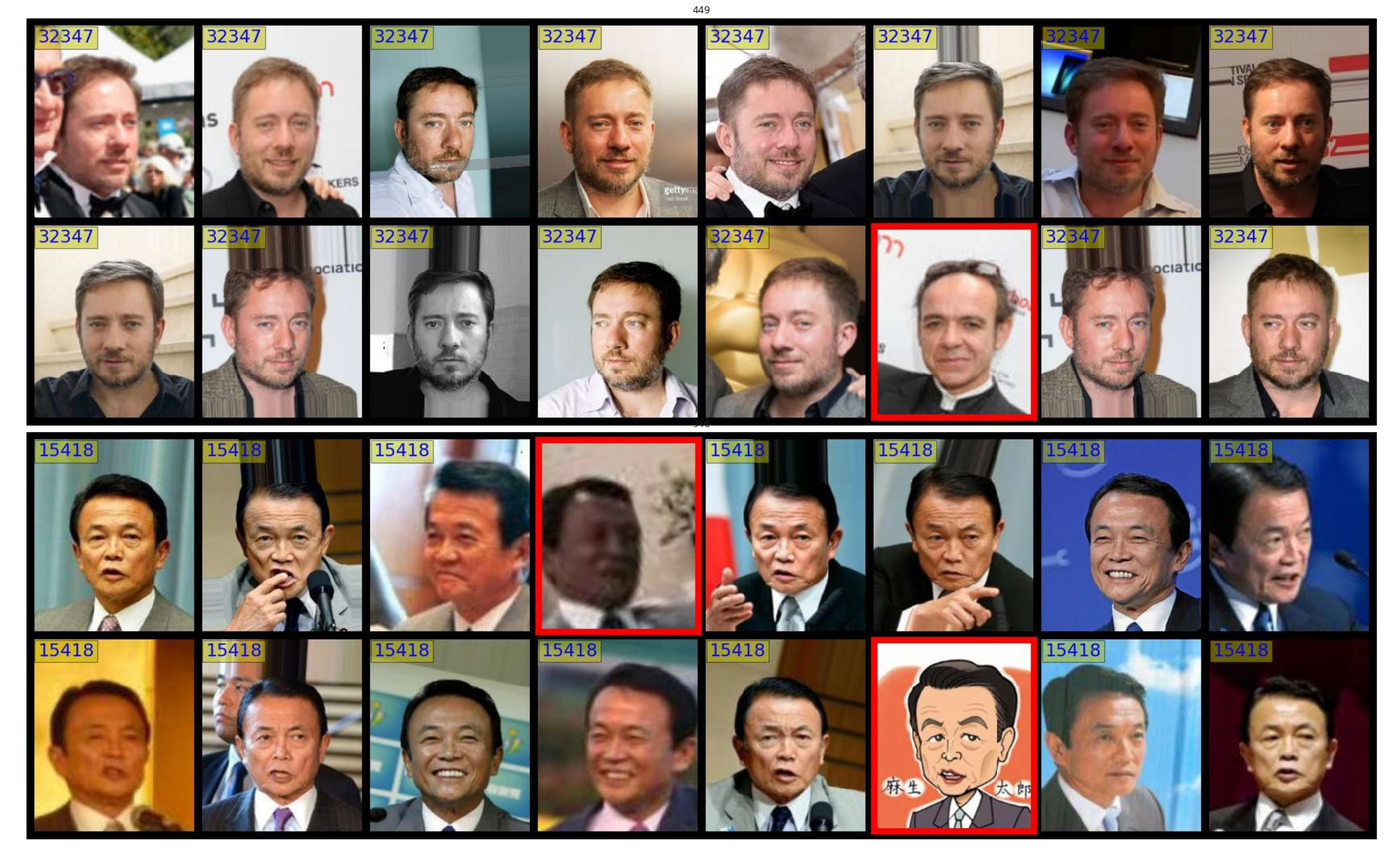}
\caption{\small{This figure shows two groups of faces in the unlabeled data. All faces in a group has the same identity according to the original annotations. The number on the top-left conner of each face is the label assigned by our proposed method, and the faces in red boxes are discarded by our method. The results suggest the high precision of our method in identifying persons of the same identity. Interestingly, our method is robust in pinpointing wrongly annotated faces (group 1), extremely low-quality faces (e.g., heavily blurred face, cartoon in group 2), which do not help training. See supplementary materials for more visual results.}}
\label{fig:vlz_denoise}
\end{figure}

\section{Conclusion}
We have proposed a novel approach, Consensus-Driven Propagation (CDP), to exploit massive unlabeled data for improving large-scale face recognition.
We achieve highly competitive results against fully-supervised counterpart by using only 9\% of the labels.
Extensive analysis on different aspects of CDP is conducted, including influences of the number of committee members, inputs to the mediator, base architecture, and committee heterogeneity.
The problem is well-solved for the first time in the literature, considering the practical and non-trivial challenges it brings.
%
%It's a promising indication that large-scale face recognition does not necessarily rely on massive annotations.

%
%We also observe that CDP can serve as a data cleaner to filter out worthless examples, which is meaningful to practical applications of face data processing.
%
%Future works include exploring optimal setting of the committee, designing better ``mediator'' for smarter pair selection, and developing the online version of CDP.

%\vspace{0.15cm}
{
\noindent
%\scriptsize
\textbf{Acknowledgement}: This work is partially supported by the Big Data Collaboration Research grant from SenseTime Group (CUHK Agreement No. TS1610626), the General Research Fund (GRF) of Hong Kong (No. 14236516, 14241716).}

\clearpage

\bibliographystyle{splncs04}
\bibliography{egbib}

\clearpage
\noindent\textbf{\large{Appendix}}
\appendix

\vspace{8pt}
\noindent
This supplementary material of paper ``Consensus-Driven Propagation in Massive Unlabeled Data for Face Recognition'' mainly includes detailed implementation of ``Consensus-Driven Propagation'' (CDP), some further analysis, and more visual results as well as the typical failure cases.

\section{Detailed Implementation}
We use PyTorch~\footnote{PyTorch: http://pytorch.org/} to implement our CNN models and the ``mediator''.
\subsection{Supervised Initialization}
For the ``tiny NASNet-A'', we use the implementation of ``NASNet-A-Large''\footnote{https://github.com/Cadene/pretrained-models.pytorch}, and keep ``x\_conv0'', ``x\_stem\_0-1'', ``x\_cell\_0-1'', ``x\_reduction\_cell\_0'', ``x\_cell\_6-7'', while removing other cells.

For all backbone architectures of the base model and the committee members, we replace the last ``average pooling'' layer by a $1\times1$ convolution layer followed by a ``fully-connected'' layer , to embed each image into a $256$ dimensional feature vector.
Then the feature vector is fed into a linear layer to produce score for each category.
The models are trained from scratch and the initialization strategy is ``Xavier''.
The batch size ranges from $256$ to $1536$, and the initial learning rate ranges from $0.5$ to $0.1$, \textit{w.r.t} different architectures.
Each batch is scattered in $8$ GPUs.
The learning rate is decayed by 10 times at epoch $\frac{2}{3.5}M$ and $\frac{3}{3.5}M$, where $M$ is the maximal number of epochs ranging from $70$ to $50$ \textit{w.r.t} different architectures.

\subsection{Consensus-Driven Propagation}
We use NMSLIB~\footnote{NMSLIB: https://github.com/searchivarius/nmslib} for cosine similarity based k-NN search, and $k$ is set to $20$ in our main comparison.

The ``mediator'' is an MLP classifier with $2$ hidden layers, each of which containing $50$ nodes.
It uses ReLU as the activation function and ``cross-entropy'' loss for binary classification.
Note that the configuration parameters of the ``mediator'' makes little difference to the final results, as long as it's not too complicated to over-fit the training pairs.
We train the ``mediator'' on $8.7M$ pairs extracted from the k-NN graph of the base model on the labeled data, for $4$ epochs until convergence.
The learning rate is $0.05$ initially, and is decayed by 10 times when epoch $3$ finishes.
In testing time, we feed pairs from unlabeled data into the trained ``mediator'' and obtain probabilities for each pair.
We set the probability threshold as $0.96$ to select high-confident pairs for the construction of the ``consensus-based graph''.

Our label propagation algorithm is shown in Algorithm~\ref{alg:propagation}.

\subsection{Joint Training}
In this stage we collect assigned labels for the unlabeled data and re-train the base model from scratch with both labeled data and unlabeled data in a multi-task manner.
The loss weights are equal to the proportion of total images in each part.
The model is trained for $10$ to $21$ epochs \textit{w.r.t} different ratios of unlabeled data ($10$ for $ratio=10$ and $21$ for $ratio=2$), and the learning rate schedule is the same as in supervised initialization.

\begin{algorithm}[t]
	\caption{Label Propagation.}
	\label{alg:propagation}
	\begin{algorithmic}[1]
		\Require find\_connected\_components: to find connected components in a graph, return list of components (sub-graphs).
		\INPUT{$\mathcal{G}$: Graph, $M$: Maximal number of nodes in each category, $Step$: Step for adjusting threshold}
		\OUTPUT{$\mathcal{RET}$: Returned nodes with labels}
		\Function{Propagation}{$\mathcal{G}, M, Step=0.1$}
		\State assert $Step < 1.0$
		\State ${\mathcal{C}_0} =$ find\_connected\_components($\mathcal{G}$)
		\State $Q = {Queue(\mathcal{C}_0)}$, $L = 0$, $\mathcal{RET} = \emptyset$
		% \State $Q = {Queue(C_0)}$
		% \State $L = 0$
		% \State $RET = \emptyset$
		\While{$Q \neq \emptyset$}
		\State $\mathcal{C} = Q.pop()$
		\If {$Size(\mathcal{C}) > M$} ~~\# partitioning by eliminating low-score edges
		\State $\mathcal{E}_{old} = \mathcal{C}.edges()$, $S_{min} = Min(\{e.score(): e \in \mathcal{E}_{old}\})$
		% \State $S_{min} = Min(\{e.score(): e \in E_{old}\})$
		%\State $S = \emptyset$
		%\For {$e = E_{old}[0]$ to $E_{old}[end]$}
		%	\State $S.append(e.score())$
		%\EndFor
		%\State $S_{min} = Min(S)$
		\State $th = S_{min} + (1 - S_{min}) \times Step$
		\State $\mathcal{E}_{new} = \{e: e \in \mathcal{E}_{old}, e.score() > th\}$
		% \State $E_{new} = \{e: e \in E_{old}, e.score() > th\}$
		%\State $E_{new} = \emptyset$
		%\For {$e \in E_{old}$}
		%	\If {$e.score() > th$}
		%		\State $E_{new}.append(e)$
		%	\EndIf
		%\EndFor
		\If {$\mathcal{E}_{new}\neq \emptyset$}
		\State $\mathcal{C} =$ find\_connected\_components(Graph($\mathcal{E}_{new}$))
		\State $Q.extend(C)$
		\EndIf
		\Else ~~\# assigning labels for eligible components
		\For {$n \in \mathcal{C}.nodes()$}
		\State $n.label = L$, $\mathcal{RET}$.append($n$)
		% \State $RET$.append($n$)
		\EndFor
		\State $L = L+1$
		\EndIf
		\EndWhile
		\State \Return $\mathcal{RET}$
		\EndFunction
	\end{algorithmic}
\end{algorithm}

\section{More Analysis}
%
\if 0
\subsection{Incorporating Advanced Loss Functions.}
%
Our CDP framework is compatible with various forms of loss functions.
%
Apart from $softmax$, we also equip CDP with an advanced loss function, ArcFace~\cite{deng2018arcface}, the current top entry on MegaFace benchmark.
%
For parameters related to ArcFace, we set the margin $m=0.5$ and adopt the output setting ``E'', that is ``BN-Dropout-FC-BN''.
%
We also use a cleaner training set aiming to obtain a higher baseline.
%
As shown in Table ~\ref{tab:arcface}, we observe that CDP still brings large improvements over this much higher baseline.
%
%\vspace{-4pt}
\begin{table}[h!]
	\centering
	%\vspace{-4pt}
	\caption{Comparisons of the gain brought by CDP with 2-folds unlabeled data between the previous baseline (Softmax) and the new baseline (ArcFace~\cite{deng2018arcface} with a cleaner training set). The performances are reported on MegaFace test set.}
	\label{tab:arcface}
	\scriptsize{
		\begin{tabular}{c|c|c}
			\hline
			& ~Softmax~			& ~ArcFace~\cite{deng2018arcface}~  \\\hline
			~baseline~   		& 61.78\%  	& 76.93\% \\\hline
			~CDP (~Ratio~=~2)~ 	& 70.51\%   & 83.68\% \\\hline
		\end{tabular}
	}
\end{table}
\fi

\subsection{One-hot Labels \textit{v.s.} Soft Labels.}
The label propagation procedure in CDP is flexible to be adapted to other label modalities.
For example, it can also propagate soft labels, \textit{i.e.}, the vector of probabilities a node belongs to each identity.
The propagation of soft labels follows an initial propagation of one-hot labels.
Then the label vectors are diffused from each node to their neighbors in breadth-first manner, with two hyper-parameters $depth$ and $decay$, standing for the maximal diffusion depth and the decay ratio of the values on each diffusion step.
Finally on each node, the values of identities are normalized to form a probability vector.
In this experiment, we adopt ``Cross Entropy Loss'' to utilize the soft labels in multi-task training.
As shown in Table~\ref{tab:softlabel}, with appropriate combination of $depth$ and $decay$, soft labels help to improve the performance on MegaFace by $1$ point, very close to the fully-supervised counterpart.
\begin{table}[h!]
	\centering
	%\vspace{-4pt}
	\caption{Comparison of propagating one-hot labels and soft labels. When $depth=0$, it is equivalent to the case of one-hot labels. In this comparison, the quantity of unlabeled data is twice the labeled data (Ratio=2).}
	\label{tab:softlabel}
	\scriptsize{
		\begin{tabular}{c|c|c|c}
			\hline
			& parameters & ~MegaFace~ & ~IJB-A~ \\
			\hline\hline
			\multirow{7}{*}{\begin{tabular}[c]{@{}c@{}}CDP\\ (Ratio=2)\end{tabular}} 
			& depth=0  			& 70.51\% 	  & 85.6\%     \\\cline{2-4}\cline{2-4}
			& depth=3, decay=0.2  & 71.21\%     & \textbf{86.70}\%    \\\cline{2-4}
			& depth=3, decay=0.5  & 70.58\%     & 85.78\%    \\\cline{2-4}
			& depth=5, decay=0.2  & 70.66\%     & 86.11\%    \\\cline{2-4}
			& depth=5, decay=0.5  & \textbf{71.48}\%     & 85.52\%    \\\cline{2-4}
			& depth=5, decay=0.8  & 70.59\%     & 85.50\%    \\\cline{2-4}
			& depth=10, decay=0.2  & 70.32\%     & 86.69\%    \\\hline\hline
			supervised	& -		& 71.5\%	  & 84.07\% \\
			\hline
		\end{tabular}
	}
\end{table}

\section{Visual Results}

\begin{figure}[t]
	\centering
	\includegraphics[width=\linewidth]{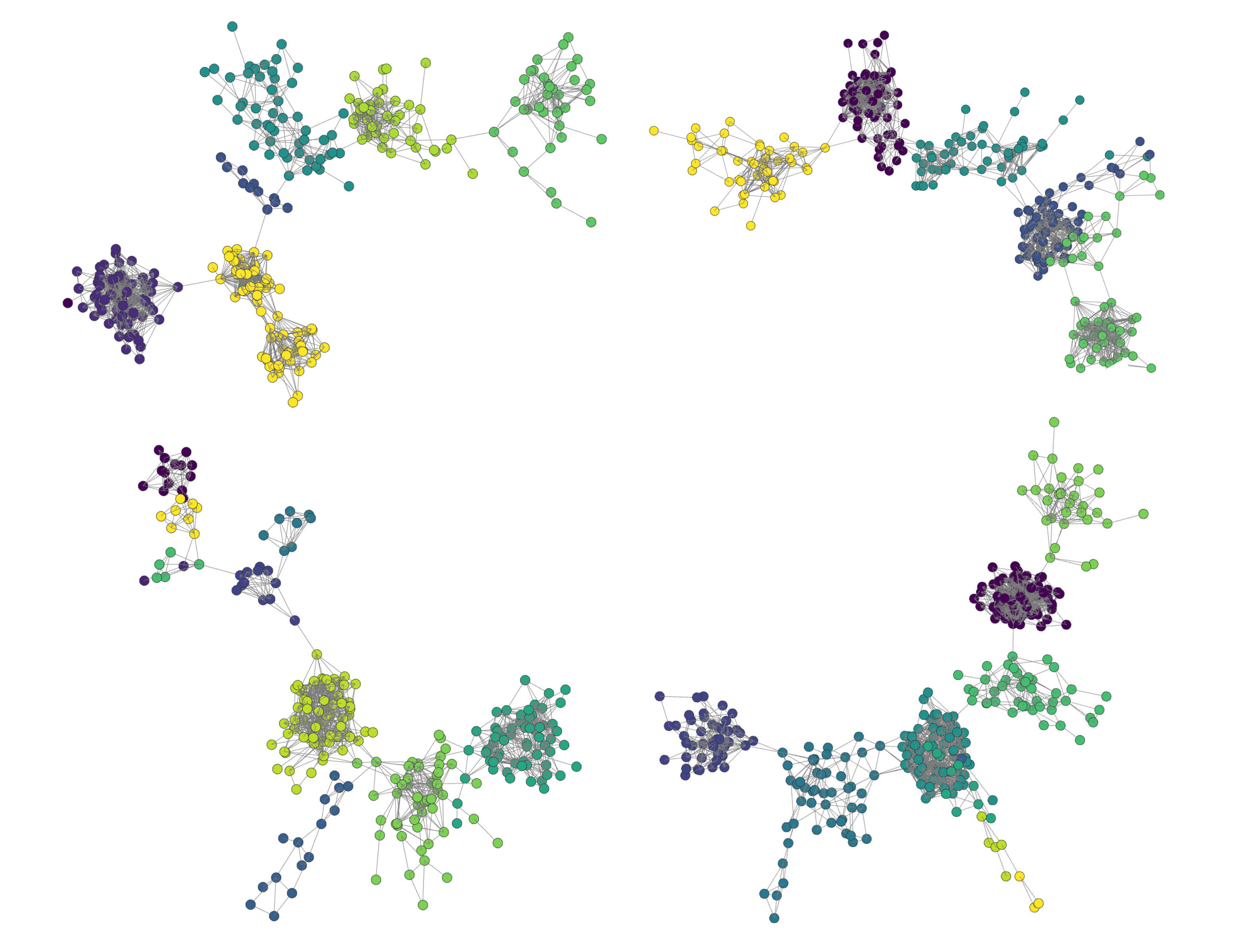}
	\caption{\small{Partial view of the ``consensus-based graph''. The colors of nodes represents the categories according to the original annotations. CDP produces dense connections among samples in the same category and weak connections between samples in different categories. Such graph facilitates the following label propagation.}}
	\label{fig:vlz_cgraph}
	%\vspace{-0.5cm}
\end{figure}

Fig.~\ref{fig:vlz_cgraph} shows partial view of the ``consensus-based graph'' in CDP.
It clearly shows that CDP produces dense connections among samples in the same category and weak connections between samples in different categories.
Such graph facilitates the following label propagation.

\begin{figure}[t]
	\centering
	\includegraphics[width=\linewidth]{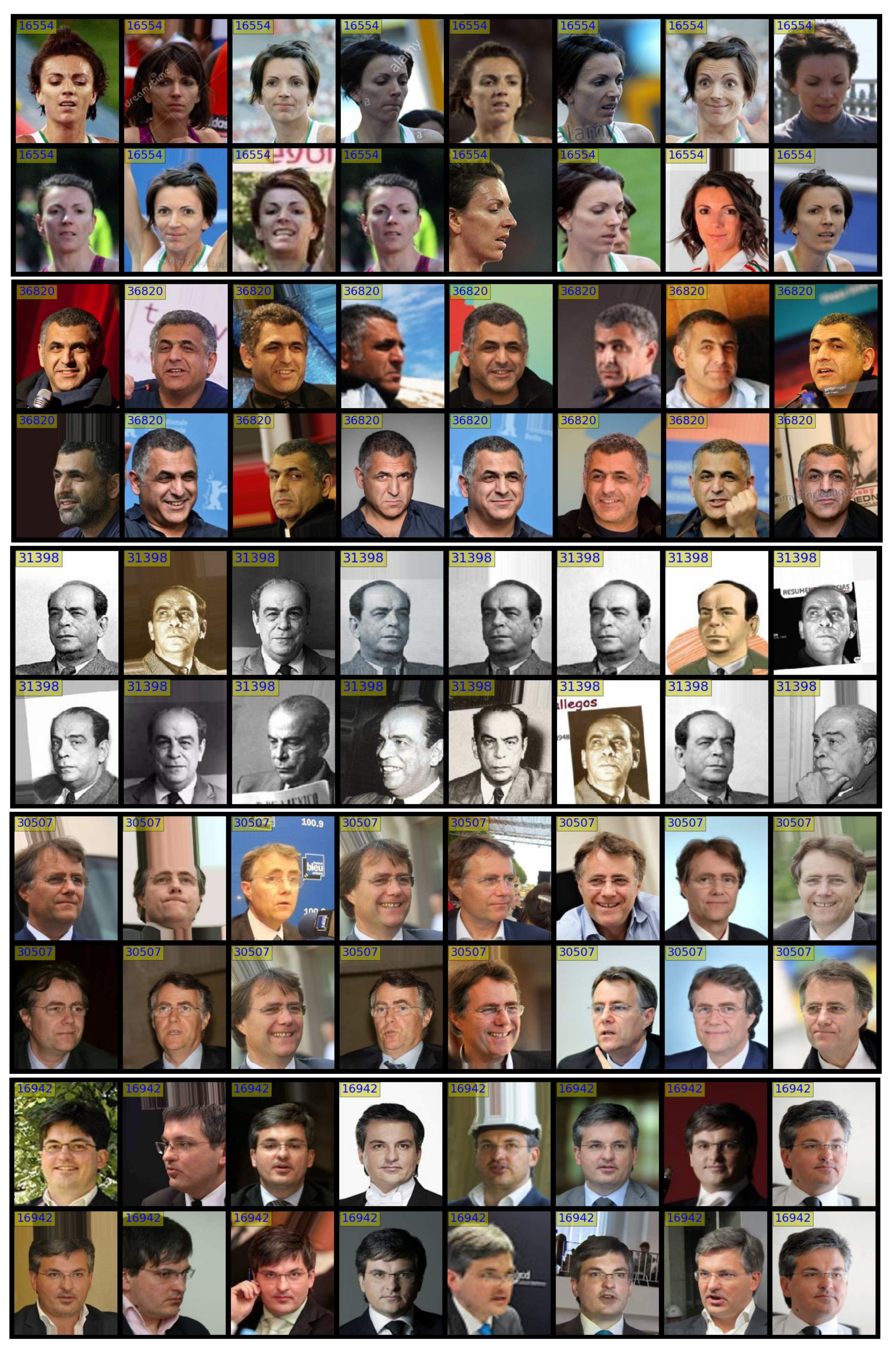}
	\caption{\small{Visual results of assigned labels by CDP. This figure shows 5 group of faces as well as their assigned labels on the top-left corner.}}
	\label{fig:vlz_group}
	%\vspace{-0.5cm}
\end{figure}

Fig.~\ref{fig:vlz_group} shows 5 groups of faces and the assigned labels.
For most of examples in the unlabeled data, CDP is able to group faces belonging to the same person together and assign the same label.

Fig.~\ref{fig:vlz_denoise} shows 4 groups of faces and CDP is able to automatically discard noisy samples.

\begin{figure}[t]
	\centering
	\includegraphics[width=\linewidth]{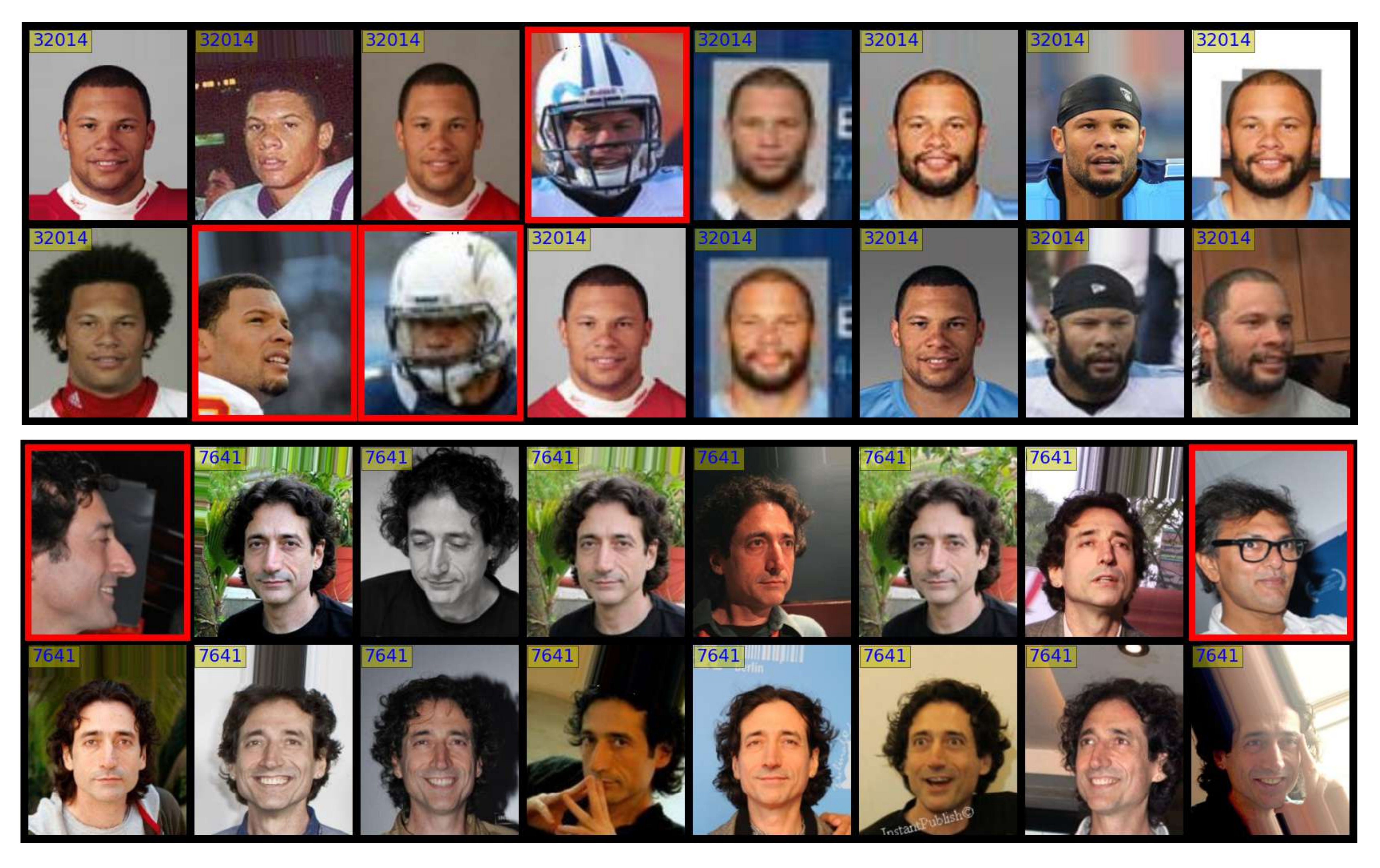}
	\caption{\small{Typical failure cases denoted by red boxes. CDP falsely discards some heavily occluded faces and atypical faces that even humans cannot easily discriminate.}}
	\label{fig:vlz_failure}
	%\vspace{-0.5cm}
\end{figure}

Fig.~\ref{fig:vlz_failure} shows the typical failure cases of our methods.
In some cases, CDP cannot identify heavily occluded faces and atypical faces that even humans cannot easily tell discriminate.
It is due to the lack of extreme training examples in the labeled data, hence both the base model and the committee trained on the labeled data cannot handle those cases well.
However, these failure cases will be handled and the performance of CDP can be continuously improved as the base model and the committee go stronger.

\begin{figure}[t]
	\centering
	\includegraphics[width=\linewidth]{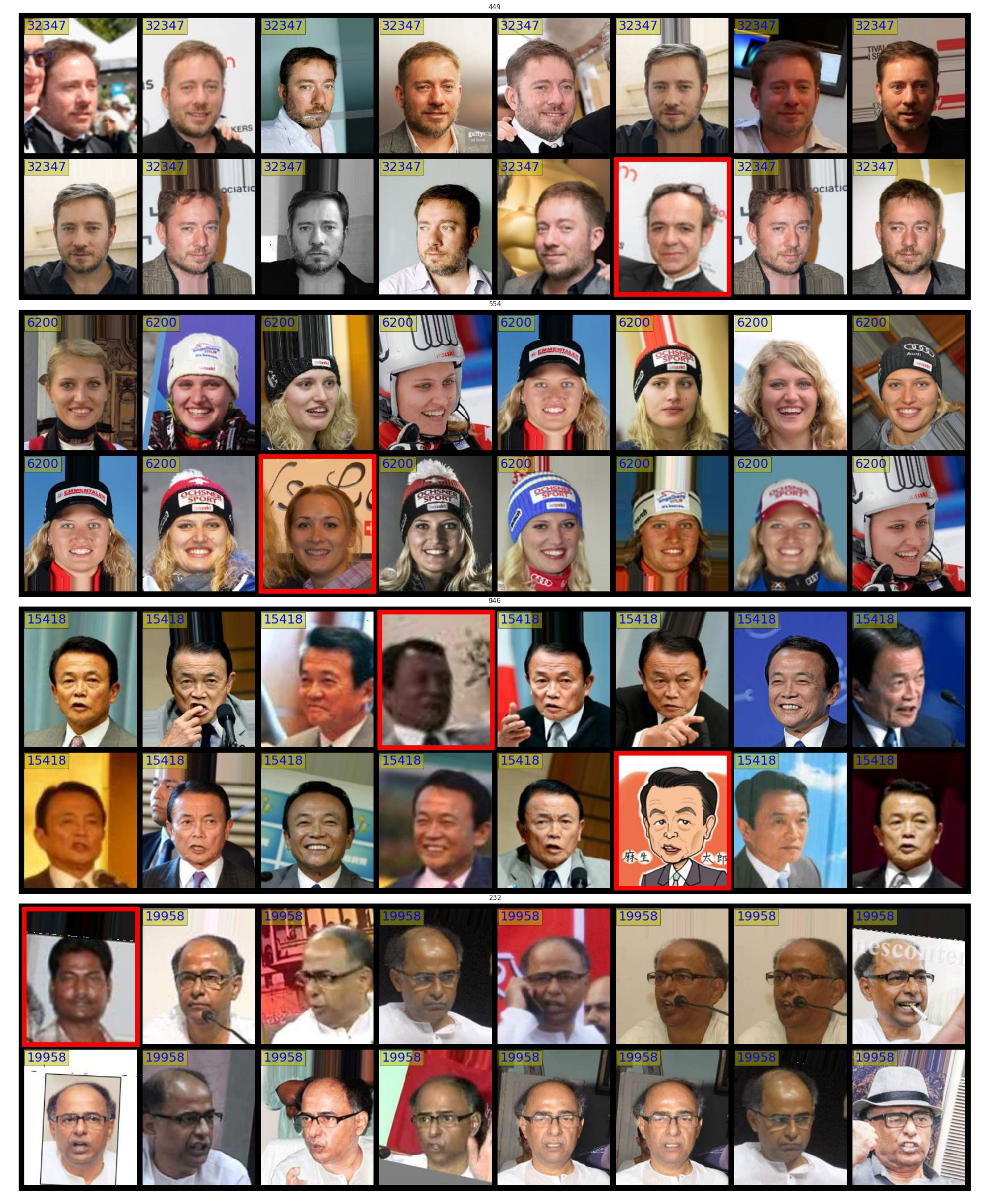}
	\caption{\small{This figure shows four groups of faces in the unlabeled data. All faces in a group has the same identity according to the original annotations. The number on the top-left conner of each face is the label assigned by our proposed method, and the faces in red boxes are discarded by our method. The results suggest the high precision of our method in identifying persons of the same identity. Interestingly, our method is robust in pinpointing wrongly annotated faces (group 1 and 2, 4), extremely low-quality faces (e.g., heavily blurred face, cartoon in group 3), which do not help training.}}
	\label{fig:vlz_denoise}
\end{figure}

\end{document}